# Idioms-Proverbs Lexicon for Modern Standard Arabic and Colloquial Sentiment Analysis

Hossam S. Ibrahim
Computer Science Department
Institute of statistical studies and research (ISSR)
Cairo University, Egypt

Sherif M. Abdou
Information Technology Department, Faculty of Computers and information
Cairo University, Egypt

Mervat Gheith
Computer Science Department
Institute of statistical studies and research (ISSR)
Cairo University, Egypt

## ABSTRACT
Although, the fair amount of works in sentiment analysis (SA) and opinion mining (OM) systems in the last decade and with respect to the performance of these systems, but it still not desired performance, especially for morphologically-Rich Language (MRL) such as Arabic, due to the complexities and challenges exist in the nature of the languages itself. One of these challenges is the detection of idioms or proverbs phrases within the writer text or comment. An idiom or proverb is a form of speech or an expression that is peculiar to itself. Grammatically, it cannot be understood from the individual meanings of its elements and can yield different sentiment when treats as separate words. Consequently, In order to facilitate the task of detection and classification of lexical phrases for automated SA systems, this paper presents AIPSeLEX a novel idioms/ proverbs sentiment lexicon for modern standard Arabic (MSA) and colloquial. AIPSeLEX is manually collected and annotated at sentence level with semantic orientation (positive or negative). The efforts of manually building and annotating the lexicon are reported. Moreover, we build a classifier that extracts idioms and proverbs, phrases from text using n-gram and similarity measure methods. Finally, several experiments were carried out on various data, including Arabic tweets and Arabic microblogs (hotel reservation, product reviews, and TV program comments) from publicly available Arabic online reviews websites (social media, blogs, forums, e-commerce web sites) to evaluate the coverage and accuracy of AIPSeLEX.

## 1. INTRODUCTION
Most of the current researches are now focusing on the area of sentiment analysis and opinion mining as the amount of internet documents increasing rapidly and where the Internet has become an area full of important information, views and meaningful discussions. The social media, blogs, forums, e-commerce web sites, etc. encourages people to share their opinion, emotions and feelings publicly which considered as very valuable information in the decision making process. These reasons make the statistical studies to evaluate the products, famous people and important topics are required and influential.

SA is the computational study of how opinions, attitudes, emotions and perspectives are expressed in text; furthermore, it contains many methods and techniques that extracting this kind of evaluative information from large datasets [1]. SA can recognize the writer's feelings as expressed in positive or negative comments, by analyzing un-readable large numbers of documents. Extensive syntactic and semantic patterns enable us to detect sentiment expressions and its orientation (sentiment polarity) which in turn used to identify whether the subjective text is positive (e.g., it is an amazing camera!), negative (e.g., I hate this cell phone) or neutral (e.g., I'll watch this film soon).

A lot of work has been conducted in the last decade to implement sentiment analysis systems in different languages with good performance and promising, such as [2-14] and others, these work are used different resources such as sentiment lexicons and tagged corpus to perform their tasks, But the need of building these resources are still ongoing, especially for morphologically-Rich language (MRL) such as Arabic, due to the complexities and challenges exists in the nature of the languages itself. One of these challenges is idioms, proverbs, saying, etc. In fact, most idioms express strong opinions and can be very informative [15], e.g., "cost an arm and a leg" which mean to be very expensive, "pinch pennies" which mean to spend as little money as possible, for Arabic idioms e.g. " أثلج صدري" "cool my breast (chest)" which mean to give relief or comfort or bring hope for good news.

Some people may prefer to be explicit and express their thoughts directly; preferring to use words and sentences in their literal sense. On the other hand, others prefer to be somewhat implicit and indirect, using idioms, sayings or proverbs leave a greater impact on the audience or maybe to capture their attention via presenting a funny, sarcast or a metaphorical expression [16]. Idioms and proverbs can be a sentence or part of a sentence (phrase), they are considered as one class of figurative expressions which occur in all expressions of at least two words which cannot be understood literally and which function as a unit semantically.

In sentiment analysis, there are two main requirements are necessary to improve sentiment analysis effectively in any language and genres; high coverage sentiment lexicons and tagged corpora to train and test the sentiment classifier [17]. Due to the lack of universal sentiment resources available for sentiment classification, in addition, human-labeled resources for each domain are costly and difficult, and the manual annotation is very expensive and time consuming. Also, to the best of our knowledge, there are no idioms/proverbs sentiment lexicons on MSA and colloquial; therefore, we present AIPSeLEX a large scale idioms/proverbs sentiment lexicon for sentiment analysis and opinion mining purposes. All the efforts that have been performed for the construction and annotation of this lexicon have been explained in details. Also the lexicon has been tested on different data, including Arabic tweets and Arabic microblogs (hotel reservation, product reviews, and TV program comments) to measure the lexicon coverage.

More than 3632 public idioms and proverbs are annotated. Although this task is time consuming, it is only a one-time effort and the annotated idioms can be used by the community. This research is concerned on MSA and Egyptian dialect where, MSA considered as the standard that commonly used by the Arab countries and Egyptian colloquial language is the most widely understood Arabic dialects [18] as will mention in the next section.

The remaining of the paper organized as follows: section2 Arabic language and challenges; section3 describes the data resources and lexicon construction of this work, also describes the techniques and methods used for detection and extraction idiom from text; section4 related the work of this area; and section5 the conclusion and our future work.

## 2. ARABIC LANGUAGE AND CHALLENGES

Arabic is one of the six official languages of the United Nations. According to Egyptian Demographic Center[1], it is the mother tongue of about 300 million people (22 countries). There are about 135 million Arabic internet users until 2013.

The orientation of writing is from right to left and the Arabic alphabet consists of 28 letters. The Arabic alphabet can be extended to ninety elements by writing additional shapes, marks, and vowels. Most Arabic words are morphologically derived from a list of roots that are tri, quad, or pent-literal. Most of these roots are tri-literal. Arabic words are classified into three main parts of speech, namely nouns, including adjectives and adverbs, verbs, and particles. In formal writing, Arabic sentences are often delimited by commas and periods. Arabic language has two main forms: Standard Arabic and Dialectal Arabic. Standard Arabic includes Classical Arabic (CA) and MSA while Dialectal Arabic includes all forms of currently spoken Arabic in daily life, including online social interaction and it vary among countries and deviate from the Standard Arabic to some extent [19]. There are six dominant dialects, namely; Egyptian, Moroccan, Levantine, Iraqi, Gulf, and Yemeni[2].

MSA considered as the standard that commonly used in books, newspapers, news broadcast, formal speeches, movies subtitles, etc. of the Arabic countries and Egyptian dialects commonly known as Egyptian colloquial language is the most widely understood Arabic dialects [18].

This research is concerned about MSA and the Egyptian dialect. The complexity here is that all the natural language processing approaches that has been applied to most languages, is not valid for applying on Arabic language directly. The text needs much manipulation and pre-processing before applying these methods. The main challenging aspects of sentiment analysis and opinion mining exist with use of other types of words, sentiment lexicon construction, dealing with negation, degrees of sentiment, complexity of the sentence/document, words in different contexts, etc.

## 3. CORPUS AND IDIOM LEXICON

### 3.1 Data collection

The data collection is built for developing and testing includes Arabic tweets, product reviews, hotel reservation comments and TV program comments. Thousands of tweets have been collected using twitter search Application Interface (API[3]) with query “lang=ar” to restrict tweets to Arabic once, also thousands of Arabic comments and reviews are collected from different microblogs and forum websites such as http://www.booking.com , http://forums.fatakat.com, http://ejabat.google.com/, etc. Then, approximately 3600 topics are randomly selected from the collected set using two native Arabic speakers. The selected data is performed according to specific conditions; Sentiment, hold one opinion, MSA, Egyptian dialect, not sarcastic, no insult, subjective and contains idiom or proverb or saying. The selected data are cleaned in a pre-processing steps such as normalization (unified Arabic characters and remove all diacritics such as ([‘أ’, ‘آ’, ‘إ’ => ‘ا’], [‘ة’, ‘ه’ => ‘ه’], [‘ي’, ‘ى’ => ‘ي’]), removing foreign characters (symbols, numbers) and reducing elongation (detecting both repeated uni-gram and bi-gram character pattern). e.g. ‘جميييييل’ (gamiiiiiiil), ‘هاهاها’ (hahaha).

### 3.2 Lexicon construction and labeling

Since this research is concerned on the Arabic language and Egyptian colloquial and most internet users who write in these languages are expressing their opinions and feelings with sarcastic way, old wisdom, old saying and idioms, which exceeded more than 20% of the total opinions (as noticed in our data). The sarcastic sentences are excluded because it is hard to deal with, e.g., “What a great car! It stopped working in two days.” Sarcasms are not so common in consumer reviews about products and services, but are very common in political discussions, which make political opinions hard to deal with [20]. But, there are a lot of phrases which represent old wisdoms and popular idioms/proverbs that people used to use it in their comments to represent their opinion directly as shown in Table1. These phrases cannot be neglected, it gives different meanings when it segmented or translated by dictionaries, therefore, there must be a glossary or lexicon covering these terms and phrases. The absence of these lexicons motivate us to build a phrase lexicon contains sentiment idioms and proverbs for MSA and Egyptian dialect.

A (32785) idioms/proverbs are collected from websites and books that specialized in this area, such as [21-25], in the time period starting from January 2014 to September 2014. We select (3632) phrases which are sentiment and commonly used, The phrases are selected and annotated manually as: positive (PO) and negative (NG) sentiment using three raters (native speakers) specialized in Arabic and linguistic knowledge of age between 30 and 40 where conflicts of tagging are resolved using the majority voting principle where the difference is discussed and a total agreement is eventually reached. The inter-annotator agreement in terms of Kappa (K) is measured. The overall observed agreement is 98%. Kappa determines the quality of a set of annotations by evaluating the agreement between annotators. The current standard metric used for measuring inter-annotator agreement on classification tasks is the Cohen Kappa statistic [26].

**Table 1. Examples of proverbs or idioms**

[1] Internet world stats “usage and population statistics”, http://www.internetworldstats.com/

[2] http://en.wikipedia.org/wiki/Varieties_of_Arabic

[3] https://search.twitter.com/

| | |
|---|---|
| **Ex1** | "تسليم السلطة للبرلمان تعني تسليم القط مفتاح الكرار" |
| | The handover of power to the parliament means delivering the key of Karar to the cat |
| **Ex2** | "بعد 30-عاما في الحكم أدركنا إن إلي افتكرناه موسي طلع فرعون" |
| | "After 30 years in power, we realized that what we thought he was Moses became Pharaoh" |

In the examples of Table1, there are not any sentiment words, and the classifiers will labeled them as neutral sentiment although they express a negative sentiment. It is clear in the phrase " تسليم القط مفتاح الكرار" of the first example which translated to "deliver the key of Karar to the cat" when translated by dictionaries in spite of it means "Give the thief the key of the safe" which expresses negative sentiment, also the phrase " افتكرناه موسي طلع فرعون" of the second example, which translated to "we thought he was Moses became Pharaoh" when translated by dictionaries, although it means, "what we thought that he was a good man, he becomes a very bad man" which expresses negative sentiment. These phrases are idioms or proverbs and must be detected and classified by the classifier using some sort of idioms lexicon such as AIPSeLEX. AIPSeLEX consists of four columns as shown in Table2: Arabic idiom, English translation, Buckwalter transliteration and sentiment polarity.

**Table 2. Examples of annotated idioms from lexicon**

| Idiom /proverb | English translation / meaning | Buckwalter | Polarity |
|---|---|---|---|
| الأطرش في الزفة | "Like the deaf in the wedding", *means having no idea about what is going on.* | Al>Tr$ fy Alzfp | NG |
| نجوم السما أقرب | "Farther than the stars in the sky", *Used to describe something that is absolutely impossible.* | njwm AlsmA >qrb | NG |
| أكل بعقله حلاوة | "He ate sweets in his mind", *manipulates with his mind and convinces him easily.* | >kl bEqlh HlAwp | NG |
| الي فات مات | "What is gone is dead" *Said it when you change your mind concerning a previous agreement or forgiveness or when a new set of rules have been put into effect.* | Aly fAt mAt | PO |
| عشم إبليس بالجنة | "Hope of the devil in heaven" *It is impossible, no way! Never think about it, it will never happen* | E$m <blys bAljnp | NG |
| الباب يفوت جمل | "The door (is big enough) for a camel". *Used to show someone that if he does not respect the rules he is not welcome.* | AlbAb yfwt jml | NG |

## 3.3 Detection and extraction

We built a technique for detecting and extracting idioms and proverbs from text to measure the coverage of AIPSeLEX and the accuracy of detecting and extracting these idioms. The algorithm consists of two steps, namely; detection step and extraction step.

**In detection step**; N-gram and lexica similarity techniques are used for idiom detection. First, all possible bi-gram to six-gram phrases are extracted from the topic as most Arabic idioms/proverbs not less than two words and measure the similarity between the extracted phrases and the idioms in the lexicon using cosine similarity [27]. Cosine similarity measure the similarity between two texts, given the value of range (1,-1), value of (1) for identical texts and (-1) for opposite texts. The similarity between two texts t1, t2 is given by:

$$\cos(\overrightarrow{t_1},\overrightarrow{t_2}) = \frac{\overrightarrow{t_1}.\overrightarrow{t_2}}{\|\overrightarrow{t_1}\|\ \|\overrightarrow{t_2}\|} \ldots\ldots\ldots\ldots\ldots\ldots\ldots\ldots.(1)$$

$$\cos(t_1,t_2) \approx sim(t_1,t_2) = \frac{\sum_{i=1}^{n} w_{i,t_1}\ w_{i,t_2}}{\sqrt{\sum_{i=1}^{n} w_{i,t_1}^2}\ .\sqrt{\sum_{i=1}^{n} w_{i,t_2}^2}} \ldots(2)$$

*Where $t_1$, $t_2$ is the two texts, n is the total number of terms in $t_1$, $t_2$, $w_{i,t1}$, $w_{i,t2}$ is the term (i) in $t_1$, $t_2$*

The detected phrase is considered as candidate idioms when its cosine similarity value is more than (0.7) which mean that more than half of its terms are similar. Actually, many experiments are performed with different thresholds till we noticed that 0.7 yield best results.

Another similarity measure method is applied which is Levenshtein distance (Edit distance) [28] to determine the correct idiom phrase from the candidate phrases and to remove noise phrases (extra phrases not an idiom and with cosine similarity more than or equal 0.7) as shown in Table3.

Levenshtein distance is a string metric for measuring the difference between two sequences of text. It is the min number of single character edits required to change one word into another using three actions (insert (Add), delete (Del), substitution (Sub)). So, the bigger the return value is, the less similar the two words are because the different words take more edits than similar one. Mathematically, the Levenshtein distance between two strings (a,b) is given by:

$$\mathbf{Lev_{a,b}=(|a|,|b|)} \quad \textit{where}$$

$$\mathrm{Lev}_{a,b}(i,j)=\begin{cases}\max(i,j) & \text{if } \min(i,j)=0,\\ \min\begin{cases}\mathrm{lev}_{a,b}(i-1,j)+1\\ \mathrm{lev}_{a,b}(i,j-1)+1\\ \mathrm{lev}_{a,b}(i-1,j-1)+1_{a_i\neq b_j}\end{cases} & \text{, \& otherwise}\end{cases}$$

*Where ($1_{a\neq b}$) is the indicator function, equal to 0 when ($a_i=b_j$) and equal 1 otherwise.*

**Topic1**: “ قلتها وأكررها المشكلة ليست في الثورة، الي ثاروا ماتوا وانما في حكم قراقوش الموجود حاليا”

**Buckwalter**: qlthA w>krrhA Alm$klp lyst fy Alvwrp, Aly vArwA mAtwA wAnmA fy Hkm qrAqw$ Almwjwd HAlyA

**Translation**: “I said it and repeat it, the problem is not in the revolution, who revolted died, but in the existing rule Qaraqush”

**Topic2**: “ انا مبكرهش حد قد ابو تريكه كلاعب لكن هو محترم طبعا في مواقفه ميدو الزملكاوي مش عاجبه تريكه انا زملكاويه وبقولك ياريتها جابت راجل ياكوتش”

**Buckwalter**: AnA mbkrh$ Hd qd Abw trykh klAEb lkn hw mHtrm TbEA fy mwAqfh mydw AlzmlkAwy m$ EAjbh trykh AnA zmlkAwyh wbqwlk yArythA jAbt rAjl yAkwt$

**Translation**: “I do not hate anyone except the football player Abu Trika but he is a respectable man, Mido Elzimlkkawi doesn’t like Trika and I’m Zimlkkawih and I say to Mido, I wish your mother had left a man”

**Table 3. Examples of errors and noises extracted using cosine SIM Techniques only**

| Data | Idioms detected | Matched idioms from lexicon | Polarity | CosineSIM Value | Is idiom? |
|---|---|---|---|---|---|
| Topic1 | حكم قراقوش<br>Hkm qrAqw$ | حكم قراقوش<br>Hkm qrAqw$ | NG | 1 | Yes |
| | الي ثاروا ماتوا<br>Aly vArwA mAtwA | الي اختشوا ماتوا | NG | 0.761 | No |
| Topic2 | وبقولك ياريتها جابت<br>wbqwlk yArythA jAbt | ياريتها جابت راجل<br>yArythA jAbt rAjl | NG | 0.812 | Noise |
| | ياريتها جابت راجل<br>yArythA jAbt rAjl | | | 1 | Yes |
| | جابت راجل ياكوتش<br>jAbt rAjl yAkwt$ | | | 0.763 | Noise |

Many experiments are performed using cosine similarity only and using Levenshtein distance only and we found that the combination of the two similarity measures yields the best results as shown in Table4, which define and detect the correct idioms/proverbs from the text and discard all wrong and noisy phrases.

**Table 4. Idioms/Proverbs detection and extraction Methods**

| Data | Detection and Extraction Techniques | Accuracy |
|---|---|---|
| Tweets, reviews and comments | N-gram+cosine similarity (only) | 81.60% |
| | N-gram+Edit Distance (only) | 86.12% |
| | N-gram+cosine similarity+ Edit Distance (combined) | 98.62% |

**In extraction step**; a heuristic rules are applied to the topic that contains idioms to prevent redundancy in the classification process of sentiment analysis systems. Specifically, the detected idioms and proverbs are replaced with text masks. For example, the idiom "crocodile tears", known to have negative (NG) sentiment polarity, should be replaced by (NG_Phrase), similarly replaced (PO_Phrase) by the idiom or proverbs that have positive sentiment. This step is very helpful in the sentiment polarity detection process of the sentiment analysis and opinion mining systems. For example, in un-supervised approaches a range of sentiment values from -3 to +3 assigned to idioms and a range of sentiment values from -1 to +1 assigned to sentiment words, etc. the net polarity of the topic can be obtained from the sign of the net score produced from all assigned values, if the net score is negative then the sentiment polarity is (NG) and vice versa. For supervised approaches it can be used in the training process, to learn the classifier that the existence of these masks increases the negativity or positivity of the topic

# 4. RELATED WORK

There has been a fair amount of work on sentiment analysis in different languages as it becomes a field of interest in the last decade. Some of the recent research has been used the idioms lexicon in their work such as: Xiaowen in [29] build an opinion mining system for customer reviews on eight (8) products using a lexicon-based approach. They used sentiment word lexicon and idiom lexicon and many features for detecting and extracting sentiment patterns from text. They manually collected more than 1000 English idioms for their lexicon. They noted that, most idioms express strong opinions although its collection and annotation is time consuming. Song and Wang in [1] built an unsupervised sentiment classifier for detecting and extracting Chinese idioms from text. They collected 24395 Chinese idioms and selected 8160 instances labeled with positive and negative classes to construct their sentiment Chinese idiom lexicon. The lexicon used to train the classifier. They evaluate their classifier and their lexicon coverage using three publicly available Chinese reviews annotated corpus of three domains (book, hotel and notebook PC).

Another sentiment analysis system (pSenti) is presented by [30]. They proposed a hybrid approach of lexicon-based approach and machine learning approach to detect the sentiment polarity and calculate the polarity strength on movie reviews and software reviews. The hybrid approach achieves high accuracy using a sentiment word lexicon and a list of 40 English idioms, 116 emoticons. They calculate the polarity strength of the text by assign a score to sentiment pattern, [-1,

+1] for sentiment word, [-2, +2] for emoticon and high score [-3, +3] for idiom, as it considered as most sentiment pattern that express high sentiment. Much of the work as mentioned that built a sentiment analysis or opinion mining systems used idioms list or lexicon to improve the sentiment classification process. However, to the best of our knowledge there is no idioms lexicon available for sentiment analysis purposes on Arabic language.

Another different work such as in [15] proved the informativity of Arabic idioms and proverbs in context. They applied a linguistic analysis of some Arabic proverbs taken from the Palestinian culture in terms of sound features, cohesion and lexical expressions showing how the uniqueness of their structure and content make them informative and interesting.

## 5. CONCLUSION AND FUTURE WORK

In this paper, we presented AIPSeLEX an idioms/proverbs sentiment lexicon of modern standard Arabic and Egyptian dialects for sentiment analysis (SA) classification tasks. We report the manual efforts that performed to build the first release of AIPSeLEX lexicon using many idioms/proverbs that are collected from different websites and books that are specialized in this area. Given the complexity, expensive and time-consuming of manual collection and annotation tasks, only 3632 idioms and proverbs are annotated. Also, the techniques and methods applied for testing the coverage and quality of the lexicon which achieved more than 90% coverage are explained. Experiments and studies proved that this lexicon will improve the sentiment classification process, and this is what already been achieved. Indeed, the lexicon is now used in Arabic sentiment analysis system (under development) and we believe it will be very useful.

For future work, we will continue in this line of research by improving our lexicon with further expansion by adding more annotated sentiment idioms, proverbs, saying and old wisdoms. Finally, this paper presents an Arabic resource that will be available to the scientific community that can be used in sentiment analysis and opinion mining purposes.